\documentclass[runningheads]{llncs}
\usepackage{graphicx}
\usepackage[]{algorithm, algpseudocode} %for Pseudo code
\usepackage{float}
\usepackage{amsmath}
\usepackage{amssymb}
\usepackage{bbm}
\usepackage{appendix}
\usepackage{cite}
\usepackage{color}
\usepackage{dirtytalk}
\usepackage{bm}

\DeclareMathOperator{\EX}{\mathbb{E}}% expected value

\DeclareMathOperator*{\argmin}{arg\,min}

\begin{document}

\title{Accounting for Gaussian Process \\ Imprecision in Bayesian Optimization\thanks{J. Rodemann would like to thank the scholarship program of Evangelisches Studienwerk Villigst for the support of his studies and Lars Kotthoff for providing data as well as Christoph Jansen and Georg Schollmeyer for valuable remarks.}}
\author{Julian Rodemann\and
Thomas Augustin \\
\email{rodemann@stat.uni-muenchen.de} \\
}
\authorrunning{J. Rodemann and T. Augustin}
\institute{Department of Statistics, Ludwig-Maximilians-Universität München (LMU)
}

\maketitle              % typeset the header of the contribution
\begin{abstract} Bayesian optimization (BO) with Gaussian processes (GP) as surrogate models is widely used to optimize analytically unknown and expensive-to-evaluate functions. In this paper, we propose Prior-mean-RObust Bayesian Optimization (PROBO) that outperforms classical BO on specific problems. First, we study the effect of the Gaussian processes' prior specifications on classical BO's convergence. We find the prior's mean parameters to have the highest influence on convergence among all prior components. In response to this result, we introduce PROBO as a generalization of BO that aims at rendering the method more robust towards prior mean parameter misspecification. This is achieved by explicitly accounting for GP imprecision via a prior near-ignorance model. At the heart of this is a novel acquisition function, the generalized lower confidence bound (GLCB). We test our approach against classical BO on a real-world problem from material science and observe PROBO to converge faster. Further experiments on multimodal and wiggly target functions confirm the superiority of our method. 

\end{abstract}

\section{Introduction}
\label{sec:BOwithGP}

Bayesian optimization (BO)\footnote{Also called efficient global optimization (EGO) or model-based optimization (MBO).} is a popular method for optimizing functions that are expensive to evaluate and do not have any analytical description (\say{black-box-functions}). Its applications range from engineering \cite{frazier2016bayesian} to drug discovery \cite{pyzer2018bayesian} and COVID-19 detection \cite{covid-AwalMHBMB21}. BO's main popularity, however, stems from machine learning, where it has become one of the predominant hyperparameter optimizers \cite{nguyen2019bayesian} after the seminal work of \cite{snoek2012practical}.

BO approximates the target function through a surrogate model. In the case of all covariates being real-valued, Gaussian Process (GP) regression is the most popular model, while random forests are usually preferred for categorical and mixed covariate spaces. BO scalarizes the surrogate model's mean and standard error estimates through a so-called acquisition function\footnote{Also referred to as infill criterion.}, that incorporates the trade-off between exploration (uncertainty reduction) and exploitation (mean optimization). The arguments of the acquisition function's minima are eventually proposed to be evaluated. Algorithm \ref{algo-bo} describes the basic procedure of Bayesian optimization applied on a problem of the sort: $\min_{\bm{x} \in \mathcal{X}} \Psi(\bm{x})$, 
    where $\Psi: \mathcal{X}^p \rightarrow \mathbb{R}$, $\mathcal{X}^p$ a $p$-dimensional covariate space. Here and henceforth, minimization is considered without loss of generality. 

\begin{algorithm}[H]
\begin{center}
\caption{Bayesian Optimization}
\label{algo-bo}
    \begin{algorithmic}[1]
    \State create an initial design $D = \{(\bm{x}^{(i)}, \Psi^{(i)})\}_{i = 1,..., n_{init}}$ of size $n_{init}$
    \While {termination criterion is not fulfilled}
    \State \textbf{train} a surrogate model (SM) on data $D$\label{algo-bo-train-sm-std}
    \State \textbf{propose} $\bm{x}^{new}$ that optimizes the acquisition function $AF(SM(\bm{x}))$ \label{algo-bo-af-std}
    \State \textbf{evaluate} $\Psi$ on $\bm{x}^{new}$ 
    \State \textbf{update} $D \leftarrow D \cup {(\bm{x}^{new}, \Psi(\bm{x}^{new}))}$
    \EndWhile
    \State \textbf{return} $\argmin_{\bm{x} \in D} \Psi(\bm{x})$ and respective $\Psi(\argmin_{\bm{x} \in D} \Psi(\bm{x}))$
    \end{algorithmic}
    \end{center}
\end{algorithm}

Notably, line \ref{algo-bo-af-std} imposes a new optimization problem, sometimes referred to as \say{auxiliary optimization}. Compared to $\Psi(\bm{x})$, however, $AF(SM(\bm{x}))$ is analytically traceable. It is a deterministic transformation of the surrogate model's mean and standard error predictions, which are given by line \ref{algo-bo-train-sm-std}. Thus, evaluations are cheap and optima can be retrieved through naive algorithms, such as grid search, random search or the slightly more advanced focus search\footnote{Focus search iteratively shrinks the search space and applies random search, see \cite[page 7]{mlrMBO}.}, all of which simply evaluate a huge number of points that lie dense in $\mathcal{X}$. Various termination criteria are conceivable with a pre-specified number of iterations being one of the most popular choices. 

As stated above, GP regressions are the most common surrogate models in Bayesian optimization for continuous covariates. The main idea of functional regression based on GPs is to specify a Gaussian process \textit{a priori} (a GP prior distribution), then observe data and eventually receive a posterior distribution over functions, from which inference is drawn, usually by mean and variance prediction. In more general terms, a GP is a stochastic process, i.e. a set of random variables, any finite collection of which has a joint normal distribution. 

\begin{definition}[Gaussian Process Regression]
\label{def:GP}
     A function $f(\bm{x})$ is generated by a \emph{Gaussian process} $\mathcal{GP}\left({m}(\bm{x}),k(\bm{x},\bm{x}')\right)$ 
    if for any finite vector of data points $(x_{1},...,x_{n})$, the associated vector of function values $\bm{f}=(f(x_{1}),...,f(x_{n}))$ has a multivariate Gaussian distribution:
  $ \bm{f} \sim \mathcal{N}\left( \bm{\mu} ,\bm{\Sigma}\right),$ where $\bm\mu$ is a mean vector and $\bm\Sigma$ a covariance matrix.   
\end{definition}

Hence, Gaussian processes are fully specified by a mean function ${m}(\bm{x}) = \EX[f(\bm{x})]$ and a kernel\footnote{Also called covariance function or kernel function.} $k_\theta(\bm{x}, \bm{x'}) = \EX\left[ \big(f(\bm{x}) - \EX[f(\bm{x})] \big) \big( f(\bm{x'}) - \EX[f(\bm{x'})] \big) \right]$ such that $f(\\bm{x})\sim\mathcal{GP}\left(m(\bm{x}),k_\theta(\bm{x},\bm{x}')\right)$, see e.g. \cite[page 13]{rasmussen2003gaussian}.
The mean function gives the trend of the functions drawn from the GP and can be regarded as the best (constant, linear, quadratic, cubic etc.) approximation of the GP functions. The kernel gives the covariance between any two function values and thus, broadly speaking, determines the function's smoothness and periodicity.

The paper at hand is structured as follows. Section~\ref{sec:sens-analysis} conducts a sensitivity analysis of classical Bayesian optimization with Gaussian processes. As we find the prior's mean parameters to be the most influential prior component, section~\ref{sec:probo} introduces PROBO, a method that is robust towards prior mean misspecification. Section~\ref{experiments} describes detailed experimental results from benchmarking PROBO to classical BO on a problem in material science. We conclude by a brief discussion of our method in section~\ref{sec:discussion}.

% Any polynomial function can serve as mean function. Any finitely positive semi-definite function (definition \ref{def: Finitely Positive Semi-Definite functions}) is a kernel function of a GP evaluated on a (finite) input vector.   

% \begin{definition}[Finitely Positive Semi-Definite functions]
% \label{def: Finitely Positive Semi-Definite functions}
% A function $f: \mathcal{X} \times \mathcal{X} \rightarrow \mathbb{R}$ is finitely positive semi-definite if
% it is symmetric ($\forall \bm{x},\bm{z} \in \mathcal{X}: f(\bm{x},\bm{z}) = f(\bm{z},\bm{x})$) and the matrix $\bm{K}$ formed by applying $f$ to any finite subset of $\mathcal{X}$ is positive semi-definite, i.e. for its quadratic form it holds $\bm{x}^T\bm{K}\bm{x} \geq 0$ $\forall \bm{x} \in \mathcal{X}$.
% \end{definition}

\section{Sensitivity Analysis}
\label{sec:sens-analysis}

\subsection{Experiments}

The question arises quite naturally how sensitive Bayesian optimization is towards the prior specification of the Gaussian process. It is a well-known fact that classical inference from GPs is sensitive with regard to prior specification in the case of small $n$. The less data, the more the inference relies on the prior information. What is more, there exist detailed empirical studies such as \cite{schmidt2008investigating} that analyze the impact of prior mean function and kernel on the posterior GP for a variety of real-world data sets. We systematically investigate to what extent this translates to BO's returned optima and convergence rates.\footnote{To the best of our knowledge, this is the very first systematic assessment of GP prior's influence on BO.} Analyzing the effect on optima and convergence rates is closely related, yet different. Both viewpoints have weaknesses: Focusing on the returned optima means conditioning the analysis on the termination criterion; considering convergence rates requires the optimizer to converge in computationally feasible time. To avoid these downsides, we analyze the mean optimizations paths. 

\begin{definition}[Mean Optimization Path]
\label{def: mop}
Given $R$ repetitions of Bayesian optimization applied on a test function $\Psi(\bm{x})$ with $T$ iterations each, let $\Psi(\bm{x}^*)_{r,t}$ be the best incumbent target value at iteration $t \in \{1, ..., T\}$ from repetition $r \in \{1, ..., R\}$. The elements 
\begin{align*} 
    MOP_t = \frac{1}{R} \sum_{r=1}^R \Psi(\bm{x}^*)_{r,t} 
\end{align*}
shall then constitute the $T$-dimensional vector $MOP$, which we call \emph{mean optimization path (MOP)} henceforth.
\end{definition}

As follows from definition \ref{def:GP}, specifying a GP prior comes down to choosing a mean function and a kernel. Both kernel and mean function are in turn determined by a functional form (e.g. linear trend and Gaussian kernel) and its parameters (e.g. intercept and slope for the linear trend and a smoothness parameter for the Gaussian kernel). Hence, we vary the GP prior with regard to the mean functional form $m(\cdot)$, the mean function parameters, the kernel functional form $k(\cdot, \cdot)$ and the kernel parameters (see definition \ref{def:GP}). We run the analysis on 50 well-established synthetic test functions from the \texttt{R} package \texttt{smoof} \cite{smoof}. The functions are selected at random, stratified across the covariate space dimensions $1, 2, 3, 4$ and $7$. For each of them, a sensitivity analysis is conducted with regard to each of the four prior components. The initial design (line 1 in algorithm \ref{algo-bo}) of size $n_{init}=10$ is randomly sampled anew for each of the $R = 40$ BO repetitions with $T=20$ iterations each. This way, we make sure the results do not depend on a specific initial sample. For each test function we obtain an accumulated difference (AD) of mean optimization paths.

\begin{definition}[Accumulated Difference of Mean Optimization Paths]
\label{def: ad}
Consider an experiment comparing $S$ different prior specifications on a test function with $R$ repetitions per specification and $T$ iterations per repetition. Let the results be stored in a $T \times S$-matrix of mean optimization paths for iterations $t \in \{1, ..., T\}$ and prior specification $s \in \{1, ...,S \}$ (e.g. constant, linear, quadratic etc. trend as mean functional form) with entries $MOP_{t,s} = \frac{1}{R} \sum_{r=1}^R \Psi(\bm{x}^*)_{r,t,s}.$
The \emph{accumulated difference (AD)} for this experiment shall then be:
\begin{align*}
AD = \sum_{t=1}^T \left( \max_s MOP_{t,s} - \min_s MOP_{t,s} \right).    
\end{align*}
\end{definition}

\subsection{Results of Sensitivity Analysis}

% The accumulated differences for all the $50$ randomly selected test functions can be found in the appendix, table \ref{tab: overall results complete}.
% , compare AD values of the Bent-Cigar function with the ones of the Matyas or the Brent function, for instance. 
The $AD$ values vary strongly across functions. This can be explained by varying levels of difficulty of the optimization problem, mainly influenced by modality and smoothness. Since we are interested in an overall, systematic assessment of the prior's influence on Bayesian optimization, we sum the $AD$ values over the stratified sample of $50$ functions. This absolute sum, however, is likely driven by some hard-to-optimize functions with generally higher $AD$ values or by the scale of the functions' target values.\footnote{Note that neither accumulated differences (definition~\ref{def: ad}) nor mean optimization paths (definition~\ref{def: mop}) are scale-invariant.} Thus, we divide each $AD$ value by the mean $AD$ of the respective function. Table~\ref{tab:results sums of ads stand.} shows the sums of these relative $AD$ values. It becomes evident that the optimization is affected the most by the functional form of the kernel and the mean parameters, while kernel parameters and the mean functional form play a minor role.

% \begin{table}[H]
%     \caption{Sum of accumulated differences of all 50 MOPs per prior specification. Comparisons between mean and kernel are more valid than between functional form and parameters.}
%     \setlength{\tabcolsep}{14pt}
%   \begin{tabular}{@{}|l|l||l|l|@{}}
%   \hline
%  Mean & Kernel & Mean & Kernel \\ 
%  functional form & functional form & parameters & parameters \\ \hline  
%   $1.8714 \cdot 10^{16}$ & $1,0876 \cdot 10^{16}$ & $1,0828 \cdot 10^{16}$ & $3.5316 \cdot 10^{15}$ \\ \hline
%     \end{tabular}

%     \label{tab:results sums of ads}
% \end{table}

\begin{table}[H]
    \caption{Sum of relative ADs of all 50 MOPs per prior specification. Comparisons between mean and kernel are more valid than between functional form and parameters.}
    \setlength{\tabcolsep}{14pt}
  \begin{tabular}{@{}|l|l||l|l|@{}}
  \hline
  Mean & Kernel & Mean & Kernel \\ 
 functional form & functional form & parameters & parameters \\ \hline  
  42.49    &            68.20                      & 77.91  &  11.40  \\ \hline
    \end{tabular}

    \label{tab:results sums of ads stand.}
\end{table}

\subsection{Discussion of Sensitivity Analysis}
Bayesian optimization typically deals with expensive-to-evaluate functions. As such functions imply the availability of few data, it comes at no surprise that the GP's predictions in BO heavily depend on the prior. Our results suggest this translates to BO's convergence. It is more sensitive towards the functional form of the kernel than towards those of the mean function and more sensitive towards the mean function's parameters than towards those of the kernel, which appear to play a negligible role in BO's convergence.     

The kernel functional form determines the flexibility of the GP and thus has a strong effect on its capacity to model the functional relationship. What is more interesting, the mean parameters' effect may not only stem from the modeling capacity but also from the optimizational nature of the algorithm. While unintended in statistical modeling, a systematic under- or overestimation may be beneficial when facing an optimization problem. Further research on interpreting the effect of the GP prior's components on BO's performance is recommended.      

\subsection{Limitations of Sensitivity Analysis}
\label{ch:sens-ana-limitations}
Albeit the random sample of $50$ test functions was drawn from a wide range of established benchmark functions, the analysis does by far not comprise all types of possible target functions, not to mention real-world optimization problems. Additionally, the presented findings regarding kernel and mean function parameters are influenced by the degree of variation, the latter being a subjective choice. Statements comparing the influence of the functional form with the parameters are thus to be treated with caution. Yet, the comparison between kernel and mean function parameters is found valid, as both have been altered by the same factors. 

What weighs more, interaction effects between the four prior components were partly left to further research. The reported $AD$ values for mean parameters and mean functional forms were computed using a Gaussian kernel. Since other kernels may interact differently with the mean function, the analysis was revisited using a power exponential kernel as well as a Matérn kernel. As we observe only small changes in $AD$ values, the sensitivity analysis can be seen as relatively robust in this regard, at least with respect to these three widely-used kernels. 
% Noteworthy, we did not control for possible interaction effects in the other direction, that is between mean functional form and parameters on the one hand with the kernel on the other hand, because the former can be (and are per default in many existing libraries) estimated from the data.\footnote{This, of course, is not possible in case of the kernel's functional form.} Hence, we measured the overall effect of kernel parameters and functional form conditioned on the best fit of the mean function.   

\section{Prior-Mean-Robust Bayesian Optimization}
\label{sec:probo}
While a highly popular hyperparameter optimizer in machine learning \cite{nguyen2019bayesian}, Bayesian optimization itself -- not without a dash of irony -- heavily depends on its hyperparameters, namely the Gaussian process prior specification. The sensitivity analysis in section~\ref{sec:sens-analysis} has shown that the algorithm's convergence is especially sensitive towards the mean function's parameters.

In light of this result, it appears desirable to mitigate BO's dependence on the prior by choosing a prior mean function that expresses a state of ignorance. Recall that Bayesian optimization is typically used for \say{black-box-functions}, where very little, if any, prior knowledge exists. The classical approach would be to specify a so-called non-informative prior over the mean parameters. However, such a prior is not unique \cite{benavoli2015prior} and choosing different priors among the set of all non-informative priors would lead to different posterior inferences \cite{mangili2016prior}. Thus, such priors cannot be regarded as fully uninformative and represent indifference rather than ignorance. Principled approaches would argue that this dilemma cannot be solved within the framework of classical precise probabilities. Methods working with sets of priors have thus attracted increasing attention, see e.g. \cite{Rios:2000, augustin2014introduction}. Truly uninformative priors, however, would entail sets of all possible probability distributions and thus lead to vacuous posterior inference. That is, prior beliefs would not change with data, which would make learning impossible. \cite{benavoli2015prior} thus propose prior \textit{near}-ignorance models as a compromise that conciliates learning and \textit{almost} non-informative priors. In the case of Gaussian processes, so-called imprecise Gaussian processes (IGP) are introduced by \cite{mangili-15-sipta} as prior near-ignorance models for GP regression. The general idea of an IGP is to incorporate the model's imprecision regarding the choice of the prior's mean function parameter, given a constant mean function and a fully specified kernel. 
In the case of univariate regression, given a base kernel $k_{\bm{\theta}}(x,x')$ and a degree of imprecision $c > 0$, \cite[definition 2]{mangili-15-sipta} defines a constant mean imprecise Gaussian process as a set of GP priors:

% We will exploit this to yield a Bayesian optimization algorithm that is more robust towards misspecifications of the prior mean parameter. 

\begin{equation}
    \mathcal{G}_{c} = \left\{ GP\left(Mh, k_{\bm{\theta}}(x,x') + \frac{1+M}{c} \right) : h = \pm 1, M \geq 0  \right\} 
 \end{equation} 
It can be shown that $\mathcal{G}_{c}$ verifies prior near-ignorance \cite[page 194]{mangili-15-sipta} and that $c \to 0$ yields the precise model \cite[page 189]{mangili-15-sipta}. Note that the mean functional form (constant) as well as both kernel functional form and its parameters do not vary in set $\mathcal{G}_c$, but only the mean parameter $ Mh$ $\in$  $]-\infty,  \infty[$. For each prior GP, a posterior GP can be inferred. This results in a set of posteriors and a corresponding set of mean estimates $\hat{\mu}(x)$, of which the upper and lower mean estimates $\underline{\hat{\mu}}(x)$, $\overline{\hat{\mu}}(x)$ can be derived analytically. To this very end, let $k_\theta(x,x')$ be a kernel function as defined in \cite{rasmussen2003gaussian}. The finitely positive semi-definite matrix $\bm{K}_n$ is then formed by applying $k_\theta(x,x')$ on the training data vector $x \in \mathcal{X}$: 
\begin{equation}
    \bm{K}_n = [k_\theta(x_i,x'_j)]_{ij}.
    \label{eq: base kernel matrix}
\end{equation}

Following \cite{mangili-15-sipta}, we call $\bm{K}_n$ base kernel matrix. Note that $\bm{K}_n$ is restricted only to be finitely positive semi-definite and not to have diagonal elements of 1. In statistical terms, $\bm{K}_n$ is a covariance matrix and not necessarily a correlation matrix. Hence, the variance $I \sigma^2$ is included. Diverging from \cite{mangili-15-sipta}, we only consider target functions without explicit noise, thus no \say{nugget term} $I \sigma^2_{nugget}$ needs to be included in $\bm{K}_n$. 

Now let $x$ be a scalar input of test data, whose $f(x)$ is to be predicted. Then $\bm{k}_x = [k_{\bm{\theta}}(x, x_1), ..., k_{\bm{\theta}}(x, x_n)]^T$ is the vector of covariances between $x$ and the training data. Furthermore, define $\bm{s}_k = \bm{K}_n^{-1} \mathbbm{1}_n $ and $\bm{S}_k = \mathbbm{1}_n^T \bm{K}_n^{-1} \mathbbm{1}_n$. Then \cite{mangili-15-sipta} shows that upper and lower bounds of the posterior predictive mean function $\hat{\mu}(x)$ for $f(x)$ can be derived. If $|\frac{\bm{s}_k \bm{y}}{\bm{S}_k}| \leq 1 + \frac{c}{\bm{S}_k}$, they are:

\begin{equation}
\overline{\hat{\mu}}(x) = \bm{k}_x^T \bm{K}_n^{-1} \bm{y} + (1 - \bm{k}_x^T \bm{s}_k) \frac{\bm{s}_k^T}{\bm{S}_k} \bm{y} + c \frac{|1-\bm{k}_x^T\bm{s}_k|}{\bm{S}_k}
\label{eq: upper mean case1}
\end{equation} 

\begin{equation}
\underline{\hat{\mu}}(x) = \bm{k}_x^T \bm{K}_n^{-1} \bm{y} + (1 - \bm{k}_x^T \bm{s}_k) \frac{\bm{s}_k^T}{\bm{S}_k} \bm{y} - c \frac{|1-\bm{k}_x^T\bm{s}_k|}{\bm{S}_k}
\label{eq: lower mean case1}
\end{equation}

If $|\frac{\bm{s}_k \bm{y}}{\bm{S}_k}| > 1 + \frac{c}{\bm{S}_k}$:

\begin{equation}
\overline{\hat{\mu}}(x) =  \bm{k}_x^T \bm{K}_n^{-1} \bm{y} + (1 - \bm{k}_x^T \bm{s}_k) \frac{\bm{s}_k^T}{\bm{S}_k} \bm{y} + c   \frac{1-\bm{k}_x^T\bm{s}_k}{\bm{S}_k}
\label{eq: upper mean case2}
\end{equation} 

\begin{equation}
\underline{\hat{\mu}}(x) = \bm{k}_x^T \bm{K}_n^{-1} \bm{y} + (1 - \bm{k}_x^T \bm{s}_k) \frac{\bm{s}_k^T \bm{y}}{c + \bm{S}_k} 
\label{eq: lower mean case2}
\end{equation} 

% \begin{definition}[Variance Estimates of IGP]
% The corresponding variance estimate of both $\overline{\hat{\mu}}(x)$ and $\underline{\hat{\mu}}(x)$ is 
% \begin{equation}
%     \widehat{\sigma}^2_{f(x)} = k_\theta(x,x) - \bm{k}_x^T \bm{K}_n^{-1} \bm{k}_x + \frac{(1 - \bm{k}_x^T \bm{s}_k)^2}{\bm{S}_k}
% \end{equation}
 
% \end{definition}

% \begin{definition}[Credible Intervals of IGP Prediction]
% For $\alpha \in [0,1]$ and $z_q$ the $q$-quantile of the standard normal distribution, the $1-\alpha$ credible interval of the mean estimate for $f(x)$ is 
% $$
% CrI_\alpha = [\underline{f}_x = \underline{\hat{\mu}}(x) - z_{1 - \frac{\alpha}{2}} \cdot \widehat{\sigma}^2_{f(x)} ,  \overline{f}_x = \underline{\hat{\mu}}(x) - z_{1 - \frac{\alpha}{2}} \cdot \widehat{\sigma}^2_{f(x)}].
% $$
% \cite[Theorem 4]{mangili-15-sipta} shows that  $CrI_\alpha = [\underline{f}_x, \overline{f}_x]$ satisfies $\overline{P}(f(x) < \underline{f}_x) \leq \frac{\alpha}{2}$ and $\overline{P}(f(x) > \overline{f}_x) \leq \frac{\alpha}{2}$.  

% \end{definition}

Inspired by multi-objective BO \cite{horn2015model}, one might think (despite knowing better) of an IGP and a GP as surrogate models for different target functions. A popular approach in multi-objective BO to proposing points based on various surrogate models is to scalarize their predictions by an acquisition function defined \textit{a priori}. The herein proposed generalized lower confidence bound (GLCB) is such an acquisition function, since it combines mean and variance predictions of a precise GP with upper and lower mean estimates of an IGP. In this way, it generalizes the popular lower confidence bound $LCB(\bm{x}) = \hat{\mu}(\bm{x}) - \tau \cdot \sqrt{var(\hat{\mu}(\bm{x}))}$, initially proposed by \cite{cox1992statistical}.\footnote{Note that from a decision-theoretic point of view, LCB violates the dominance principle. GLCB inherits this property.}

% The only difference from GLCB to a multi-objective acquisition function stems from the fact that the different surrogate models are estimated with regard to one and the same (single-objective) target function instead of multiple ones.

\begin{definition}[Generalized Lower Confidence Bound (GLCB)]
\label{def: glcb}
Let $\bm{x} \in \mathcal{X}$. As above, let $\overline{\hat{\mu}}(\bm{x}),  \underline{\hat{\mu}}(\bm{x})$ be the upper/lower mean estimates of an IGP with imprecision $c$. Let $\hat{\mu}(\bm{x})$ and $var(\hat{\mu}(\bm{x}))$ be the mean and variance predictions of a precise GP. The prior-mean-robust acquisition function \emph{generalized lower confidence bound (GLCB)} shall then be \begin{gather*}
    GLCB(\bm{x}) = \hat{\mu}(\bm{x}) - \tau \cdot \sqrt{var(\hat{\mu}(\bm{x}))} - \rho \cdot (\overline{\hat{\mu}}(\bm{x}) - \underline{\hat{\mu}}(\bm{x})).
\end{gather*}
\end{definition}

By explicitly accounting for the prior-induced imprecision, GLCB generalizes the trade-off between exploration and exploitation: $\tau > 0 $ controls the classical \say{mean vs. data uncertainty} trade-off (degree of risk aversion) and $\rho > 0 $ controls the \say{mean vs. model imprecision} trade-off (degree of ambiguity aversion). Notably, $\overline{\hat{\mu}}(\bm{x}) - \underline{\hat{\mu}}(\bm{x})$ simplifies to an expression only dependent on the kernel vector between $x$ and the training data $\bm{k}_x = [k_{\bm{\theta}}(x, x_1), ..., k_{\bm{\theta}}(x, x_n)]^T$, the base kernel matrix $\bm{K}_n$ (equation~\ref{eq: base kernel matrix}) and the degree of imprecision $c$, which follows from equations~\ref{eq: upper mean case2} and \ref{eq: lower mean case2} in case $|\frac{\bm{s}_k \bm{y}}{\bm{S}_k}| > 1 + \frac{c}{\bm{S}_k}$:

\begin{equation}
    \overline{\hat{\mu}}(x) -  \underline{\hat{\mu}}(x) = (1 - \bm{k}_x^T \bm{s}_k) \big( \frac{\bm{s}_k^T}{\bm{S}_k} \bm{y} + \frac{c}{\bm{S}_k} - \frac{\bm{s}_k^T \bm{y}}{c + \bm{S}_k}  \big)
    \end{equation}

As can be seen by comparing equations~\ref{eq: upper mean case1} and~\ref{eq: lower mean case1}, in case of $|\frac{\bm{s}_k \bm{y}}{\bm{S}_k}| \leq 1 + \frac{c}{\bm{S}_k}$, the model imprecision $\overline{\hat{\mu}}(\bm{x}) - \underline{\hat{\mu}}(\bm{x})$ even simplifies further: $\overline{\hat{\mu}}(x) - \underline{\hat{\mu}}(x) =  2 c \frac{|1-\bm{k}_x^T\bm{s}_k|}{\bm{S}_k}$. In this case, the GLCB comes down to $GLCB(\bm{x}) = \hat{\mu}(\bm{x}) - \tau \cdot \sqrt{var(\hat{\mu}(\bm{x}))} -  2 \cdot \rho c \frac{|1-\bm{k}_x^T\bm{s}_k|}{\bm{S}_k}$ and the two hyperparameters $\rho$ and $c$ collapse to one. In both cases, the surrogate models $\underline{\hat{\mu}}(x)$ and $\overline{\hat{\mu}}(x)$ do not have to be fully implemented. Only $\bm{K}_n$ and $\bm{k}_x = [k_{\bm{\theta}}(x, x_1), ..., k_{\bm{\theta}}(x, x_n)]^T$ need to be computed. GLCB can thus be plugged into standard BO without much additional computational cost.\footnote{Further note that with expensive target functions to optimize, the computational costs of surrogate models and acquisition functions in BO can be regarded as negligible.} Algorithm \ref{algo-glcb} describes the procedure.

\begin{algorithm}[H]
\begin{center}
\caption{Prior-mean-RObust Bayesian Optimization (PROBO)}
\label{algo-glcb}
    \begin{algorithmic}[1]
    \State create an initial design $D = \{(\bm{x}^{(i)}, \Psi^{(i)})\}_{i = 1,..., n_{init}}$ of size $n_{init}$
    \State specify $c$ and $\rho$
    \While {termination criterion is not fulfilled}
    \State \textbf{train} a precise GP on data $D$ \label{algo-bo-train-sm} and obtain $\hat{\mu}(\bm{x})$, $var(\hat{\mu}(\bm{x}))$
    \State \textbf{compute} $\bm{K}_n$ and $\bm{k}_x = [k_{\bm{\theta}}(x, x_1), ..., k_{\bm{\theta}}(x, x_n)]^T$
    
    \State  \textbf{if} $|\frac{\bm{s}_k \bm{y}}{\bm{S}_k}| > 1 + \frac{c}{\bm{S}_k}$ \textbf{then}
    \State{ $\overline{\hat{\mu}}(x) - \underline{\hat{\mu}}(x) = (1 - \bm{k}_x^T \bm{s}_k) \big( \frac{\bm{s}_k^T}{\bm{S}_k} \bm{y} + \frac{c}{\bm{S}_k} - \frac{\bm{s}_k^T \bm{y}}{c + \bm{S}_k}  \big)$}  
    \State \textbf{else}  $\overline{\hat{\mu}}(x) - \underline{\hat{\mu}}(x) =  2 c \frac{|1-\bm{k}_x^T\bm{s}_k|}{\bm{S}_k}$ 
    
    \State \textbf{compute} $GLCB(\bm{x}) = - \hat{\mu}(\bm{x}) + \tau \cdot \sqrt{var(\hat{\mu}(\bm{x}))} + \rho \cdot (\overline{\hat{\mu}}(\bm{x}) - \underline{\hat{\mu}}(\bm{x}))$
    \State \textbf{propose} $\bm{x}^{new}$ that optimizes $GLCB(\bm{x})$
    \State \textbf{evaluate} $\Psi$ on $\bm{x}^{new}$ 
    \State \textbf{update} $D \leftarrow D \cup {(\bm{x}^{new}, \Psi(\bm{x}^{new}))}$
    \EndWhile
    \State \textbf{return} $\argmin_{\bm{x} \in D} \Psi(\bm{x})$ and respective $\Psi(\argmin_{\bm{x} \in D} \Psi(\bm{x}))$
    \end{algorithmic}
    \end{center}
\end{algorithm}

Just like LCB, the generalized LCB balances optimization of $\hat{\mu}(x)$ and reduction of uncertainty with regard to the model's prediction variation $\sqrt{var(\hat{\mu}(\bm{x}))}$ through $\tau$. What is more, GLCB aims at reducing model imprecision caused by the prior specification, controllable by $\rho$. Ideally, this would allow returning optima that are robust not only towards classical prediction uncertainty but also towards imprecision of the specified model. 

\section{Results}
\label{experiments}

\begin{figure}
    \centering
        \includegraphics[scale = 0.45, trim={0 0 12cm 0},clip]{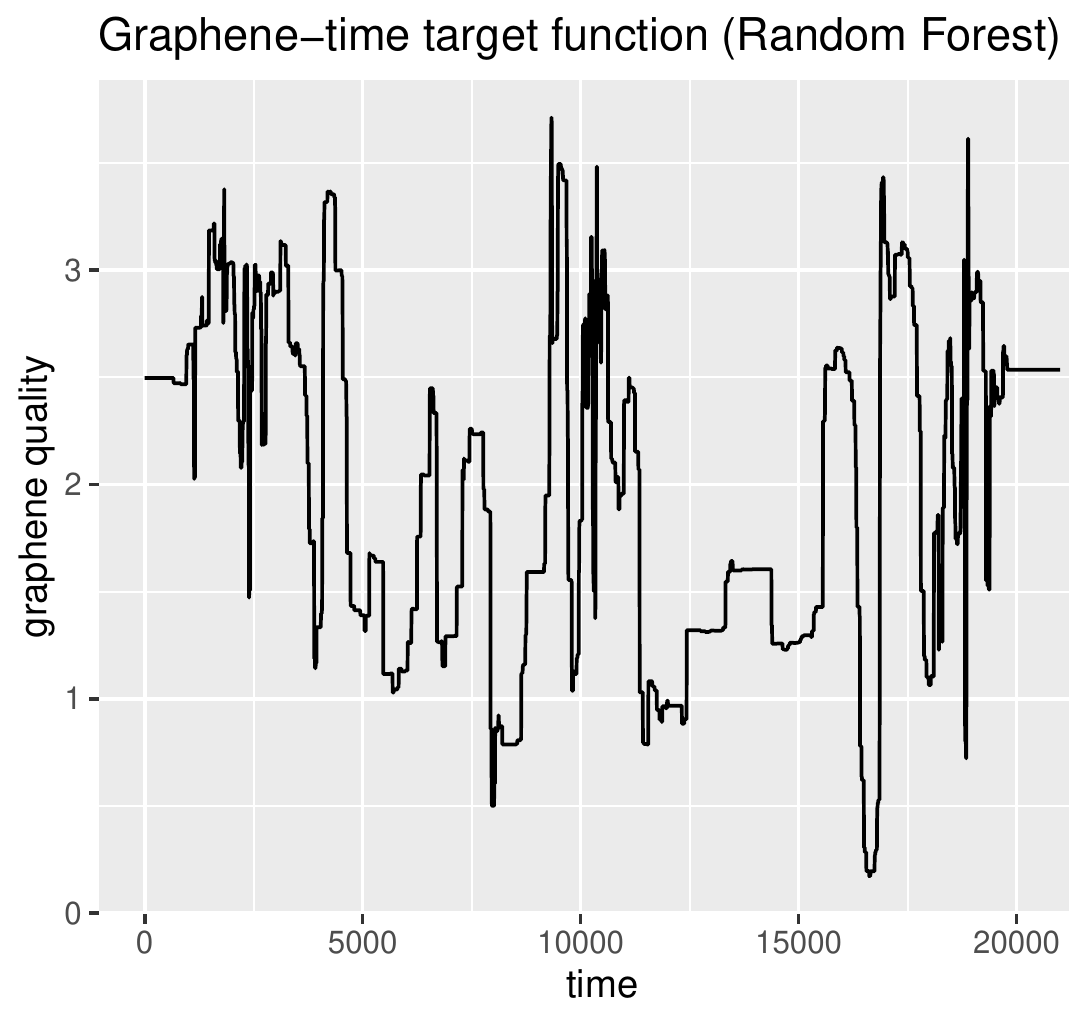}
    \label{fig:laser}\\
    \caption[Graphene Quality Functions]{Univariate target function estimated from graphene data.}
    \label{fig:graphene-funs}
\end{figure}

We test our method on a univariate target function generated from a data set that describes the quality of experimentally produced graphene, an allotrope of carbon with potential use in semiconductors, smartphones and electric batteries \cite{wahab2020machine}. The data set comprises $n = 210$ observations of an experimental manufacturing process of graphene. A polyimide film, typically Kapton, is irradiated with laser in a reaction chamber in order to trigger a chemical reaction that results in graphene. Four covariates influence the manufacturing process, namely power and time of the laser irradiation as well as gas in and pressure of the reaction chamber \cite{wahab2020machine}. The target variable (to be maximized) is a measure for the quality of the induced graphene, ranging from $0.1$ to $5.5$. 
In order to construct a univariate target function from the data set, a random forest was trained on a subset of it (target quality and time, see figure \ref{fig:graphene-funs}). The predictions of this random forests were then used as target function to be optimized.  
  
% \begin{table}[H]
%         \caption[Graphene Data]{Graphene data set \cite{wahab2020machine}.}   

% \centering
%         \begin{small}
%         \begin{tabular}{|c |c |c |c |c |c}
%           \hline
%          feature & min & max & type & description\\
%           \hline
%         power & 10 & 5555 & real-valued & power of the laser\\
%         time & 500 & 20210 & real-valued & irradiation time\\
%         gas &  &  & categorical & gas used in the reaction chamber\\ 
%         & & & & (Nitrogen, Argon, Air)\\
%         pressure & 0 & 1000 & real-valued & pressure in the reaction chamber\\

%         target quality & 0.1 &5.5 & real-valued & quality of induced graphene\\
%           \hline
%         \end{tabular}
%         \end{small}
%         \label{tab:graphene}
% \end{table}

\begin{figure}
    \centering
    \includegraphics[scale=0.46]{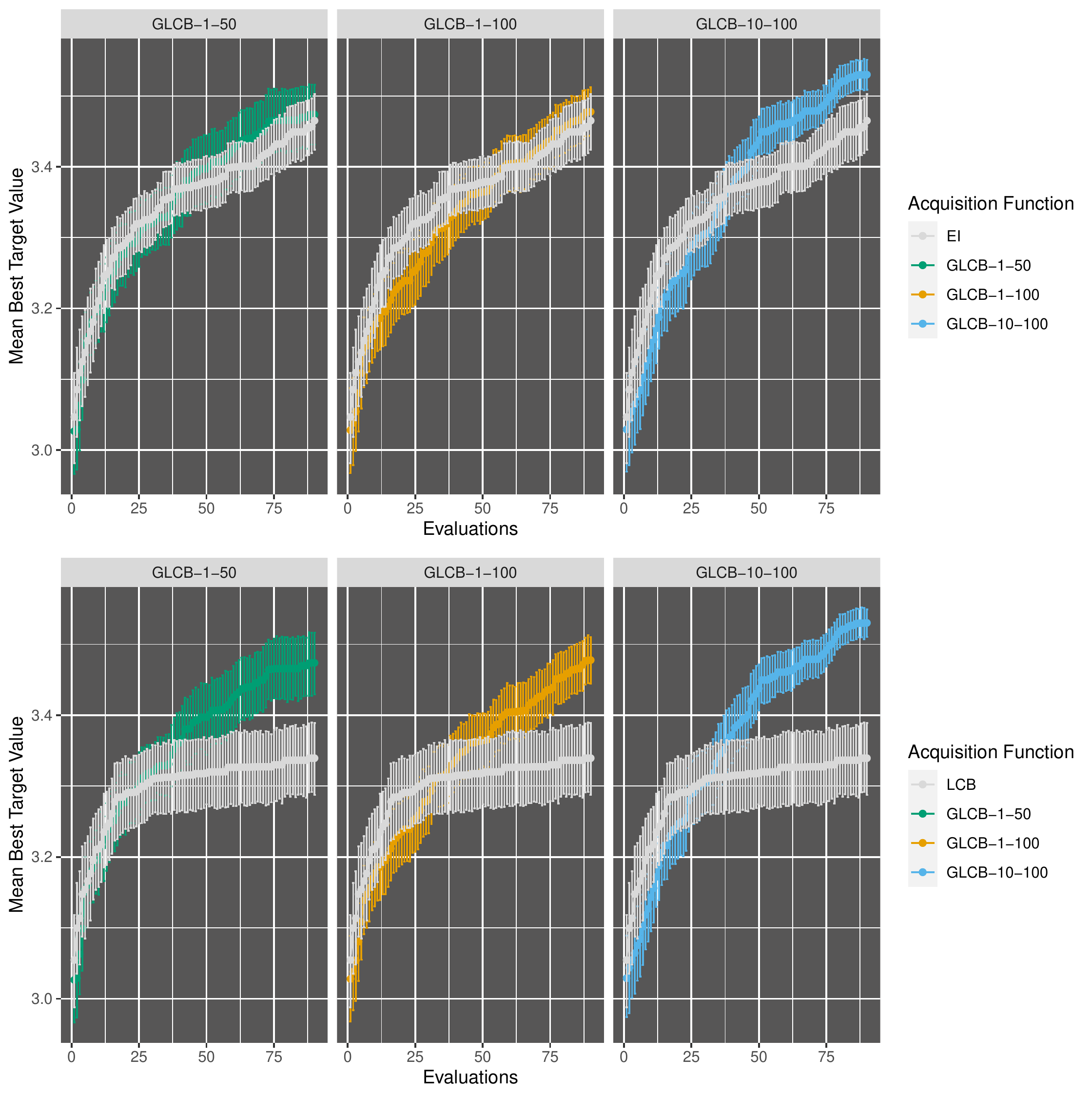}
    \caption[GLCB vs. LCB and EI on Graphene Data]{Benchmarking results from graphene data: Generalized lower confidence bound (GLCB) vs. expected improvement (EI) and lower confidence bound (LCB). Shown are 60 runs per Acquisition Function with 90 evaluations and initial sample size 10 each. Error bars represent 0.95-Confidence Intervals. GLCB-1-100 means $\rho$ = 1 and c~=~100; $\tau$ = 1 for all GLCBs and LCB. Code to reproduce results is available on this repository: \url{https://github.com/rodemann/gp-imprecision-in-bo}.}
    \label{fig:glcb-vs-lcb-and-ei}
\end{figure}

We compare GLCB to its classical counterpart LCB as well as to the expected improvement (EI), which is usually considered the most popular acquisition function. It was initially proposed by \cite[pages 1-2]{mockus1975bayesian}, disguised as a utility function in a decision problem that captures the expected deviation from the extremum. Let $\psi(\bm{x})$ be the surrogate model, in our case the posterior predictive GP, and $\Psi_{min}$ the incumbent minimal function value. The expected improvement at point $\bm{x}$ then is 
$EI(\bm{x}) = \EX( \max\{\Psi_{min} - \psi(\bm{x}), 0\} ).$ For pairwise comparisons of GLCB to LCB and EI, we observe $n = 60$ BO runs with a budget of 90 evaluations and an initial design of $10$ data points generated by latin hypercube sampling \cite{mckay2000comparison} each. Focus search \cite[page 7]{mlrMBO} was used as infill optimizer with 1000 evaluations per round and 5 maximal restarts. All experiments were conducted in \texttt{R} version 4.0.3 \cite{R} on a high performance computing cluster using 20 cores (linux gnu). Figure \ref{fig:glcb-vs-lcb-and-ei} depicts mean optimization paths of BO with GLCB compared to LCB and EI on the graphene-time target function. The paths are shown for three different GLCB settings: $\rho = 1, c = 50$ and $\rho = 1, c = 100$ as well as $\rho = 10, c = 100$. Figure \ref{fig:glcb-vs-lcb-and-ei} shows that GLCB surpasses LCB (all settings) and EI ($\rho=10, c=100$) in late iterations. We also compare GLCB to other acquisition functions and retrieve similar results, except for one purely exploratory and thus degenerated acquisition function, see chapter 6.7.2 in \cite{rodemann2021robust}.

Further benchmark experiments are conducted on meteorological data, heartbeat time series as well as synthetic functions from \cite{smoof}. In case of multimodal and wiggly target functions, the results resemble figure \ref{fig:glcb-vs-lcb-and-ei}. When optimizing smooth functions, however, classical acquisition functions like EI and LCB are superior to GLCB. For a detailed documentation of these further experiments, we refer the reader to chapters 6.6.2, 6.6.3 and 6.6.4 in \cite{rodemann2021robust}.

\section{Discussion}
\label{sec:discussion}

%\subsection{Limitations}
The promising results should not hide the fact that the proposed modification makes the optimizer robust only with regard to possible misspecification of the mean function parameter given a constant trend. Albeit the sensitivity analysis conducted in section~\ref{sec:sens-analysis} demonstrated its importance, the mean parameter is clearly not the only influential component of the GP prior in BO. For instance, the functional form of the kernel also plays a major role, see table \ref{tab:results sums of ads stand.}. The question of how to specify this prior component is discussed in \cite{malkomes2018automating, duvenaud2014automatic}. Apart from this, it is important to note that PROBO depends on a subjectively specified degree of imprecision $c$. It does not account for any imaginable prior mean (the model would become vacuous, see section~\ref{sec:probo}). What is more, it may be difficult to interpret $c$ and thus specify it in practical applications. However, our method still offers more generality than a precise choice of the mean parameter. Specifying $c$ corresponds to a weaker assumption than setting precise mean parameters. Notwithstanding such deliberations concerning PROBO's robustness and generality, the method simply converges faster than BO when faced with multimodal and non-smooth target functions. The latter make up an arguably considerable part of problems in hyperparameter-tuning and engineering.

%\subsection{Outlook}

The herein proposed method opens several venues for future work. An extension to other Bayesian surrogate models seems feasible, since there is a variety of prior near-ignorance models. What is more, also non-Bayesian surrogate models like random forest can be altered such that they account for imprecision in their assumptions, see \cite{utkin2019imprecise} for instance. Generally speaking, imprecise probability (IP) models appear very fruitful in the context of optimization based on surrogate models. They not only offer a vivid framework to represent prior ignorance, as demonstrated in this very paper, but may also be beneficial in applications where prior knowledge is abundant. In such situations, in the case of data contradicting the prior, precise probabilities often fail to adequately represent uncertainty, whereas IP models can handle these prior-data conflicts, see e.g. \cite{walter2009imprecision}.

\appendix

\bibliographystyle{splncs04}
\bibliography{literature.bib}





% LNCS Homepage, \url{http://www.springer.com/lncs}. Last accessed 4
% \end{thebibliography}

\end{document}